\documentclass{article}

\usepackage{PRIMEarxiv}

\usepackage[utf8]{inputenc} 
\usepackage[T1]{fontenc}    
\usepackage{hyperref}       
\usepackage{url}            
\usepackage{booktabs}       
\usepackage{amsfonts}       
\usepackage{nicefrac}       
\usepackage{microtype}      
\usepackage{lipsum}
\usepackage{fancyhdr}       
\usepackage{graphicx}       
\graphicspath{{media/}}     
\usepackage{subcaption}
\title{The Effect of Balancing Methods on Model Behavior in Imbalanced Classification Problems}

\author{\\
	\href{https://orcid.org/0009-0006-8819-0268}{\includegraphics[scale=0.06]{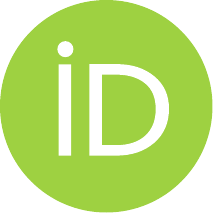}\hspace{1mm}Adrian Stando} \\
	Warsaw University of Technology \\
	Faculty of Mathematics and Information Science \\
	Warsaw, Poland \\
	\texttt{adrian.stando.stud@pw.edu.pl} \\
        \\
        \href{https://orcid.org/0000-0002-6172-5449}{\includegraphics[scale=0.06]{orcid.pdf}\hspace{1mm}Mustafa~Cavus} \\
	Eskisehir Technical University \\
	Department of Statistics \\
	Eskisehir, Turkey \\
	\texttt{mustafacavus@eskisehir.edu.tr} \\
	\\
	\href{https://orcid.org/0000-0001-8423-1823}{\includegraphics[scale=0.06]{orcid.pdf}\hspace{1mm}Przemysław~Biecek} \\
	Warsaw University of Technology \\
	Faculty of Mathematics and Information Science \\
	Warsaw, Poland \\
	\texttt{przemyslaw.biecek@pw.edu.pl} \\
}

\begin{document}
\maketitle

\begin{abstract}
Imbalanced data poses a significant challenge in classification as model performance is affected by insufficient learning from minority classes. Balancing methods are often used to address this problem. However, such techniques can lead to problems such as overfitting or loss of information. This study addresses a more challenging aspect of balancing methods - their impact on model behavior. To capture these changes, Explainable Artificial Intelligence tools are used to compare models trained on datasets before and after balancing. In addition to the variable importance method, this study uses the partial dependence profile and accumulated local effects techniques. Real and simulated datasets are tested, and an open-source Python package \texttt{edgaro} is developed to facilitate this analysis. The results obtained show significant changes in model behavior due to balancing methods, which can lead to biased models toward a balanced distribution. These findings confirm that balancing analysis  should go beyond model performance comparisons to achieve higher reliability of machine learning models. Therefore, we propose a new method \texttt{performance gain plot} for informed data balancing strategy to make an optimal selection of balancing method by analyzing the measure of change in model behavior versus performance gain.
\end{abstract}

\keywords{Imbalanced learning \and Explainable artificial intelligence \and Data balancing \and Model behavior change \and Performance gain plot}

\section{Introduction}
Classification is one of the most common machine learning (ML) tasks, providing solutions in a wide variety of fields. It frequently involves imbalanced target variables, where there is only one class of particular importance, but there are much fewer data examples available for that class than for the other. This is very common in real-world applications such as credit score prediction, heart attack risk assessment, and fraud detection. However, it can be challenging to train models on such data because many ML models assume a uniform distribution of target variables. If this is not satisfied, the algorithms may lose their ability to learn from the data. One of the approaches to deal with this problem is to apply balancing methods. They are based on undersampling and oversampling, or they combine both approaches. 

Although there are numerous balancing methods proposed in the literature, none of them is universally superior. Each has its advantages and disadvantages, and the most appropriate one depends on the specific characteristics of the dataset and the task at hand. For example, oversampling techniques cause excessive learning of the model, which can lead to overfitting, while undersampling methods cause loss of information. Even though such problems, posed by many methods, have been examined from various perspectives, very few studies have examined how they affect how the models behave.

The field that focuses on the study of model behavior is Explainable Artificial Intelligence (XAI). It provides tools that help make the decisions made by models comprehensible and transparent to humans. Consequently, it is possible to understand how balancing methods change the predictions made by the model. This can be done by examining and measuring the extent of changes in the explanations of models trained on the original and balanced datasets. So far, the researchers have only examined these changes by comparing the variable importance (VI) of the models.
VI tools provide information about the order of importance of the variables in the model but do not provide any information about the change in the relationships between the variables. In this study, we investigate the effects of the balancing methods on the model behavior using the partial dependence profiles, which determine the relationships between the response variable and the explanatory variables. 
In addition, we also use accumulated local effect profiles, which are more robust to correlated features and may provide a more accurate representation of model behavior. To measure the extent of change in model explanations, we developed a novel metric called SDD. We used it to compare models trained with logistic regression, random forest, and gradient boosting algorithms on both simulated and real unbalanced datasets. This research uses two different types of data to provide a more controlled assessment of changes in model behavior. This is supported by the fact that after applying balancing methods to real-world datasets, it is difficult to determine which one represents the true ground truth because the model structure has changed. To address this issue, we also perform simulations using synthetic datasets where the ground truth is known. We propose the performance gain plot that can be used for the selection of the optimal balancing method to solve the dilemma that arises due to the negative effects of balancing methods on model behavior as well as improving model prediction performance. In addition, to facilitate the evaluation of different balancing methods and XAI tools, we have developed a Python package that provides a unified interface for data preprocessing, model training, and XAI analysis. The package includes implementations of several popular balancing methods and XAI tools, as well as customized evaluation metrics to measure the impact of balancing methods on model behavior.

The main contributions of this paper are as follows: (1) to investigate the effects of balancing methods on model behavior, (2) to propose a measure to quantify changes in model behavior, (3) to create benchmark datasets with imbalanced class distributions, (4) to propose the performance gain plot for optimal balancing method selection regarding the performance gain versus the model behavior change, and (5) to develop a Python package that simplifies the workflow of using balancing methods, training models, and applying XAI tools. In the remainder of this paper, we first discuss the related works in Sec.~\ref{sec:related_works}, then we present the XAI tools used in the experiments and the proposed comparison metric in Sec.~\ref{sec:method}, describe our experiments conducted on simulated and real datasets in Sec.~\ref{sec:experiments}, and discuss the results and conclusions in Sec.~\ref{sec:conclusions}.

\section{Related Works} \label{sec:related_works}
The focus of this paper is to explore how XAI tools offer a new perspective on the problem of imbalanced learning. In this section, we provide an overview of existing methods for addressing the issue of imbalanced learning and a summary of research on their impact on model behavior. In addition, we emphasize the distinctive contributions of our study compared to similar work in the field.

In many domains where binary classification is applied, the class of interest is extremely rare. Classification models tend to favor the majority class in such cases, leading to bias. This bias results in a higher frequency of misclassification of minority class examples. The problem of bias towards the majority class has been addressed through several proposed methods, which can be divided into two groups: algorithmic-level methods \cite{Gu_et_al_2022, Li_et_al_2022}, which aim to develop better algorithms, and data-level methods \cite{Chawla_et_al_2002, He_et_al_2008}, which involve transforming the original dataset to balance it.

There are many studies focused on how data balancing influences and changes the performance of trained ML models \cite{Vazquez_et_al_2023, Gu_et_al_2022}. However, there have not been many attempts to investigate how they affect the model's behavior. \cite{Patil_et_al_2020} investigated the changes in the order of importance of the variables in the model after applying the balancing methods. They studied whether the balancing technique SMOTE changes the correlations between features. The experiment was conducted only on one highly imbalanced dataset. The results show that this algorithm was successful not only in eliminating imbalance but also in preserving the original correlations. The authors then applied a few XAI methods to extract the explanations of the model trained on oversampled data. However, they emphasized that it could be done only because the feature correlations remained unchanged. 
\cite{Alarab_and_Prakoonwit_2022} sought an answer to a similar question in an application on blockchain data. They compared the explanations of models trained on the dataset after applying different balancing methods. The feature importance is used and compared with a statistical test. The experiments were conducted on two datasets using different variants of the SMOTE algorithm. The results show that one of the methods changed the feature importance in both cases. However, these studies have two main limitations: (1) they do not provide enough comprehensive information about the change in model behavior as they only use feature importance, and (2) their results cannot be generalized because they are based on only one or two datasets. Moreover, \cite{Saarela_and_Jauhiainen_2021} showed that the most important features differ depending on the variable importance technique used. They suggested using a combination of the explanation techniques could provide more consistent results.

In this paper, we aim to investigate the effects of balancing methods on model behavior which is firstly mentioned in \cite{cavus_and_biecek_2022}. To do so, we propose a new metric based on the differences between the partial dependence profile and the accumulated local effect profiles since it is not possible to directly measure the model behavior over the PDP and ALE profiles by using the existing metrics proposed in \cite{Schwalbe_and_Finzel_2023, Visani_et_al_2022, Roy_et_al_2022, Zhang_et_al_2022, Agarwal_et_al_2022}.

\section{Methodology} \label{sec:method}
This section presents the XAI tools used and the proposed metric for measuring the change in model behavior.

\subsection{Partial dependence profile}
The partial dependence profile (PDP) is introduced in \cite{Friedman_2001}. Let $X$ be the data set and $X^j$ be any variable in the data set. The PDP is a function of the observation $z$ for a model $f$ and a variable $j$ defined as follows:

\begin{equation}
    PDP(f, j, z) = E_{X^{-j}}[f(X^{j|=z})].
\end{equation}

In other words, the PDP value for the $j$-th column in the observation $z$ is an average prediction of model $f$ when values in the $j$-th column are set to $z$. In practice, however, we do not usually know the distribution of $X^{-j}$ \cite{ema}. Therefore, it is estimated using the following formula: 

\begin{equation}
    \widehat{PDP}(f, j, z) = \frac{1}{n} \sum_{i=1}^n f(X_i^{j|=z}).
\end{equation}

\subsection{Accumulated local effects}
PDP may provide explanations that can be misleading if explanatory features are correlated. Therefore, the Accumulated Local Effects (ALE) profiles are proposed \cite{apley_and_zhu_2020}. It is a noteworthy alternative to PDP because both produce the functions as outputs, but ALE is unbiased. ALE for a model $f$ and a variable $j$ is a function of observation $z$ defined as follows:

\begin{equation}
    ALE(f, j, z) = \int_{z_0}^z \Big( E_{X^{-j}} \Big[ {\frac{\partial f(u)}{\partial u^j}}_{u = X^{j|=v}} \Big] \Big) dv + c.
\end{equation}

\noindent The constant $c$ is selected in such a way that $E_{X^j}[ALE(f, j, X^j)] = 0$ and $z_0$ is a value close to the lower bound of the support of $X^j$. In other words, ${\frac{\partial f(u)}{\partial u^j}}$ describes the local change of the model which is then averaged over the distribution of $X^{-j}$ and integrated over values from $z_0$ to $z$. The estimator of ALE is the following:

\begin{equation}
    \widehat{ALE}(f, j, z) = \sum_{k = 1}^{K} \left( \frac{1}{\sum_l w_l^j (z_k)} \sum_{i = 1}^N w_i^j (z_k) \left[f\left(x^{j|=z_k}\right) - f\left(X^{j|=z_k - \Delta}\right)\right] \right) + \hat{c}.
\end{equation}

\noindent The constant $K$ is the number of intermediate points $(z_0, z_1, ..., z_K)$ are evenly distributed points in $(z_0, z)$ interval with step $\Delta = (z - z_0) / K$. The weights $w_i^j(z_k)$ represent the distance between $z_k$ and $x_i^j$. The constant $\hat{c}$ is selected in such a way that $\sum_{i = 1}^{N} ALE(f, j, X_i^j) = 0$ \cite{ema}.

\subsection{Variable importance}
The previous two methods show how the model predictions change when the value of the selected variable changes. The Variable Importance (VI) tool, however, focuses on creating one explanation for all variables in the model. We use the VI which permutes each column in $X$ multiple times and see how it affects the model performance - the method proposed by \cite{Fisher_et_al_2019}.

\subsection{Standard Deviation of the Differences (SDD)}
Assume that $f_1$ and $f_2$ be two models trained on the same data set, the SDD metric:
\begin{equation}
    SDD(h_1, h_2, j, k) = sd[{h_1(x_i) - h_2(x_i)}_{i = 1, 2, ..., k}]
\end{equation}

\noindent where $h_1(z) = PDP(f_1, j, z)$ and $h_2(z) = PDP(f_2, j, z)$ are the values of the profiles for the $j$-th variable at point $z$. In the above definition, if $h_1(z) = h_2(z), \forall z$, which means that the profiles are equal, the metric value $SDD(h_1, h_2, j, k) = 0$. Similarly, if $h_1(z) = h_2(z) + c, \exists c \forall z$, which means that the profiles are parallel and $SDD(h_1, h_2, j, k) = 0$. This behavior is expected as the metric is intended to measure the changes in the shape of the curves, not the vertical offset. This is because the position of the PDP curve depends on the accuracy of the model, but only the changes in shape indicate changes in behavior.

\begin{figure}
    \begin{subfigure}{0.48\textwidth}
      \includegraphics[width=\linewidth]{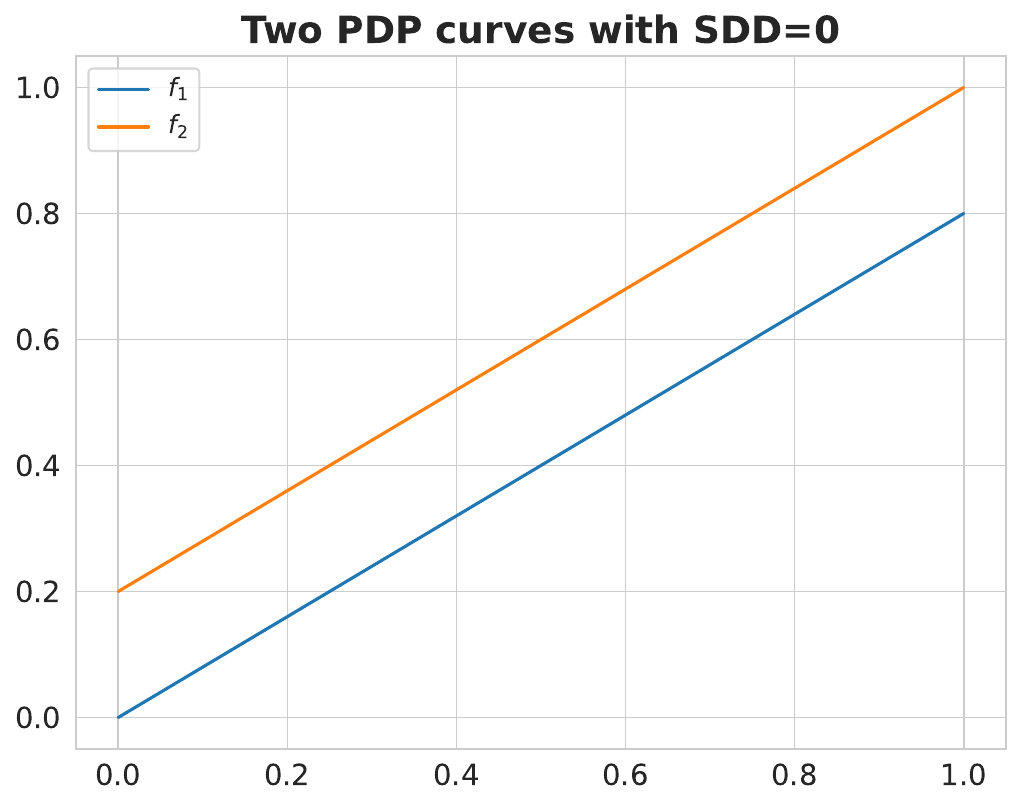}
      \caption{Two example PDP curves with $SDD=0$.}
      \label{fig:sdd-1}
    \end{subfigure}\hfill
    \begin{subfigure}{0.48\textwidth}
      \includegraphics[width=\linewidth]{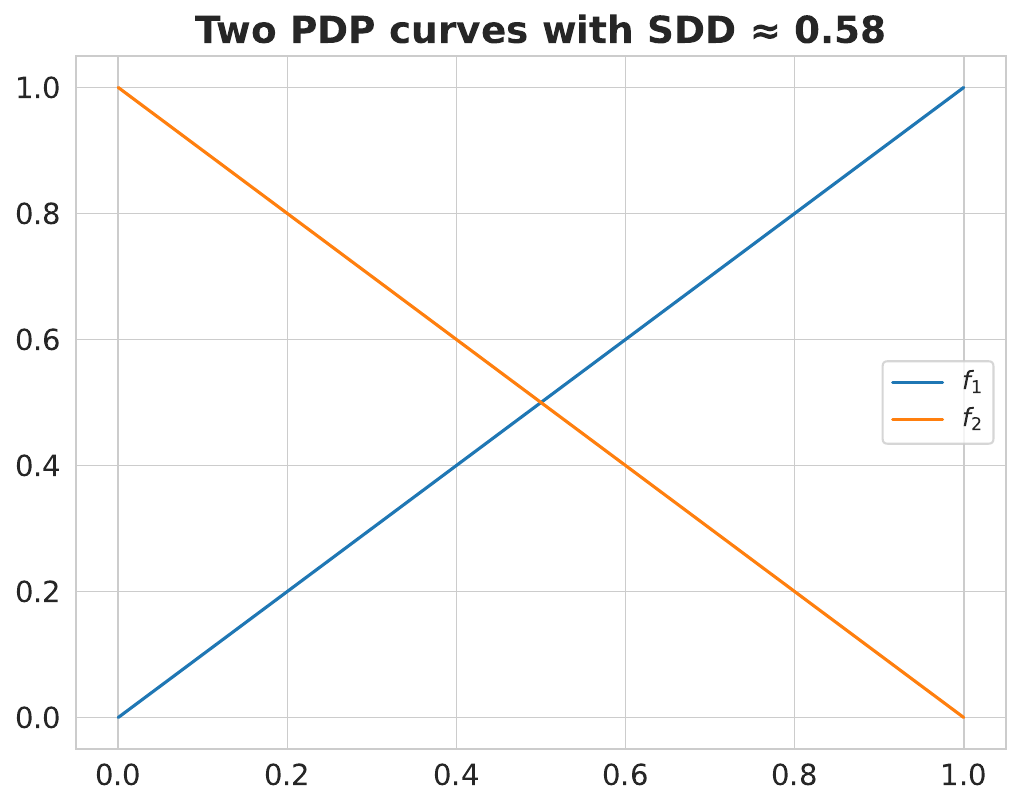}
      \caption{Two example $PDP$ curves with $SDD \approx 0.58$.}
      \label{fig:sdd-2}
    \end{subfigure}
    \caption{Exemplary $PDP$ plots and the $SDD$ values.}
    \label{fig:sdd-examples}
\end{figure}

On the other hand, consider $(x_1, x_2, ..., x_{101}) = (\frac{0}{100}, \frac{2}{100}, ..., \frac{100}{100})$, $h_1(z) = z$, and $h_2(z) = 1 - z$. In such a situation, $SDD(h_1, h_2, j, 101) \approx 0.58$. As can be seen in Figure~\ref{fig:sdd-examples}, the curves that behave differently have a high value of $SDD$. It compares two models for one variable. 
Additionally, the $SDD$ values can be aggregated to the averaged $SDD$ ($ASDD$) values to compare the behavior of models with respect to all variables. This can be formalized as follows: 
\begin{equation}
 ASDD(f_1, f_2, k) = \frac{1}{m} \sum_{j=1}^{m} SDD[PDP(f_1, j, *), PDP(f_2, j, *), j, k].   
\end{equation}

\section{Experiments} \label{sec:experiments}
In this section, we conduct experiments to measure the impact of six balancing methods: Undersampling (Random, Near Miss), Oversampling (Random, SMOTE, Borderline SMOTE), and Hybrid (SMOTETomek)  on the behavior of the three models (Logistic Regression, Random Forest, and XGBoost) behaviors on simulated and real imbalanced datasets in terms of SDD of PDP and ALE. The \texttt{edgaro} Python package (Explainable imbalanceD learninG compARatOr) is implemented to run the experiments. It is the first to provide a user-friendly interface to balance and train ML models for several datasets arranged in arrays or nested arrays. It allows using implementations from two major libraries, \textit{scikit-learn} and \textit{imbalanced-learn}. The package also calculates, in the same form of arrays and nested arrays, explanations using the PDP, ALE, or VI method and provides functions to compare them. To ensure that the explanations for each experiment are calculated over the same data, the test dataset extracted before any balancing is used as the background data. By default, the test size is equal to 20\%, and the data is split in a stratified fashion (preserving the class distribution in both subsets).

\subsection{Simulated Dataset Experiments}

We conducted experiments on simulated data, which were generated through simulations that allowed us to control the ground truth \cite{Amiri_et_al_2020}. We were unsure which model represented the ground truth, as there are changes in the model behavior due to the use of balancing methods. For this purpose, we constructed a comprehensive model framework and followed a simulation design similar to \cite{Casalicchio_et_al_2019}, and we used the following model to simulate the imbalanced datasets:

\begin{equation}
    z = \beta_{0i}+ 2.9 X_1 - 3.7 X_2 + 1.2 X_3 + \epsilon_j
\end{equation}

\noindent where the binary response variable $Y \sim B(1, p = 1 / (1 + exp(-z)))$, the explanatory variables $X_1, X_2, X_3 \sim N(0, 1)$ and the error term $\epsilon_j \sim N(0, v_j)$. The simulations are set up as 12 scenarios: $\beta_{0i}$ takes the values in $\{1.5, 2.5, 3.5, 4.5\}$ and the variance of the error term $v_j$ takes the values in $\{1, 2, 3\}$ respectively, in the scenario $ij$ to generate the dataset. The error term $\epsilon$ controls the variance of the model prediction, and the parameter $\beta_{0i}$ controls the imbalance ratio of the target variable in the dataset.

\begin{figure}
   \centering
   \includegraphics[scale = 0.5]{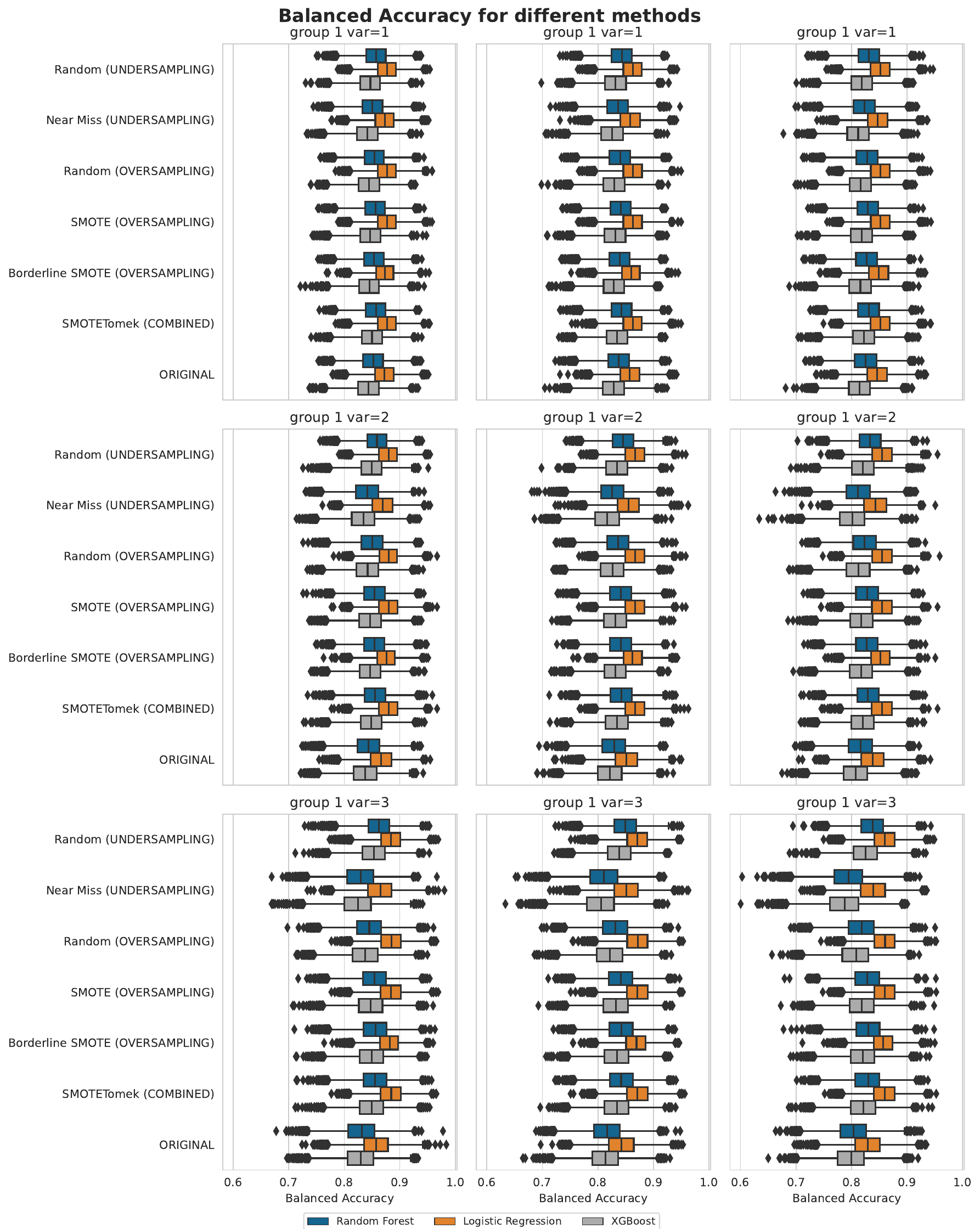}
   \caption{Balanced Accuracy of the models trained on the simulated dataset. \texttt{group} represents the $\beta_0$ values (group $i$: $\beta_{0i} = \{1.5, 2.5, 3.5, 4.5\}$), and \texttt{var} represents the variance of the error term ($var_j = \{1, 2, 3\}$)}
   \label{fig:truth-methods}
\end{figure}

Figure~\ref{fig:truth-methods} shows how the balancing methods improve the model performance in terms of balanced accuracy. The plots at the bottom of each panel show the distribution of the model performance, which is trained on imbalanced datasets as the reference level. As the value of the coefficient $\beta_0$ increases, indicating an increase in the imbalance ratio, the performance of the model decreases. Similarly, an increase in the variance of the error term leads to a decrease in the model's performance. When evaluated separately, logistic regression, random forests, and XGBoost exhibit the highest performance, respectively. Among them, increasing the variance of the error term has the largest impact on reducing the performance of the random forest model. When examining the positive effects of balancing methods on model performance, Random Undersampling is the most effective method, followed by all Oversampling methods. The Near Miss method reduces the performance of the random forest and XGBoost models as the imbalance ratio increases.

\begin{figure}
    \centering
    \includegraphics[scale = 0.5]{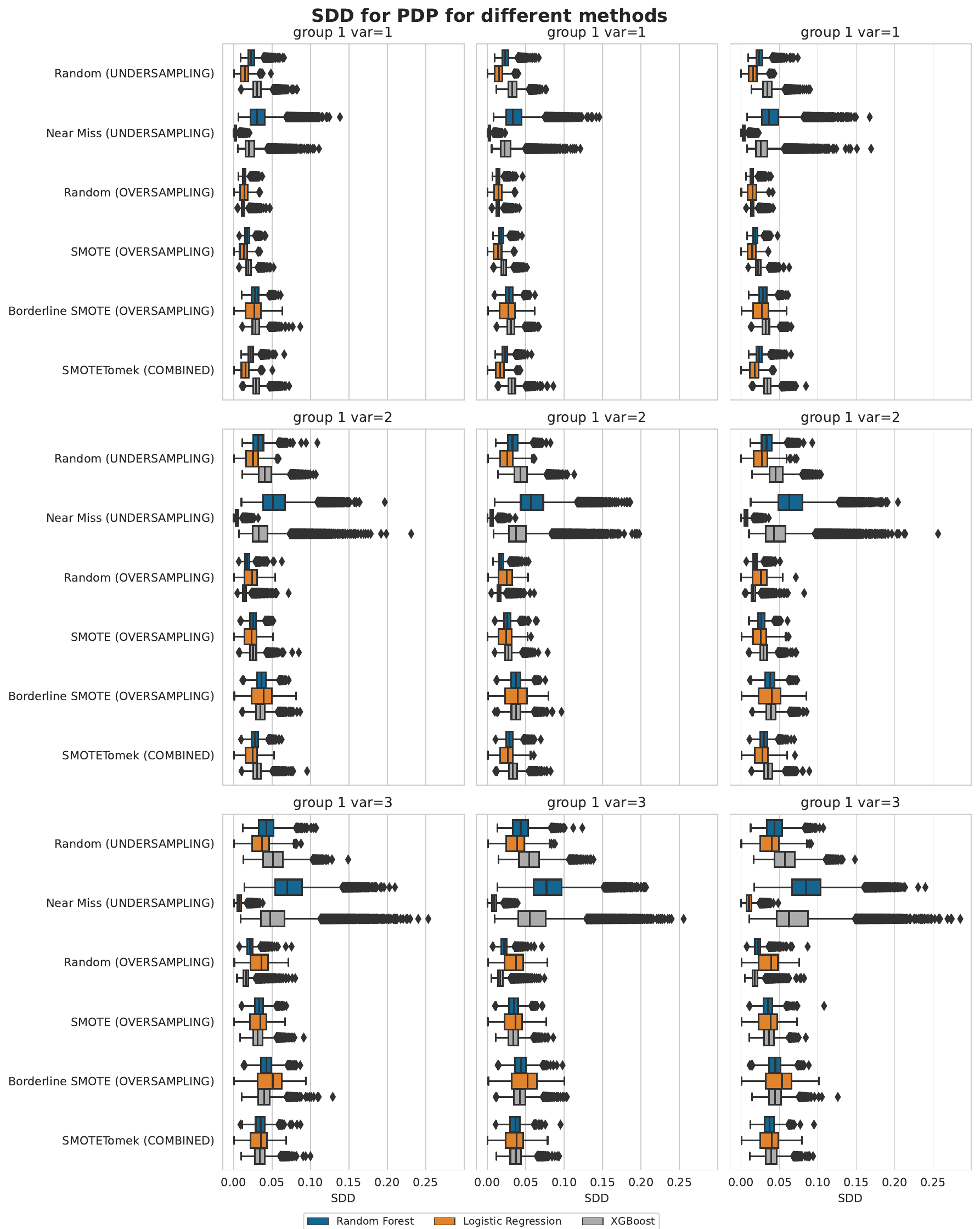}
    \caption{SDD results based on partial dependence profiles of the models trained on the simulated dataset. \texttt{group} represents the $\beta_0$ values (group $i$: $\beta_{0i} = \{1.5, 2.5, 3.5, 4.5\}$), and \texttt{var} represents the variance of the error term ($var_j = \{1, 2, 3\}$)}
    \label{fig:truth-pdp}
\end{figure}

\begin{figure}
    \centering
\includegraphics[scale = 0.5]{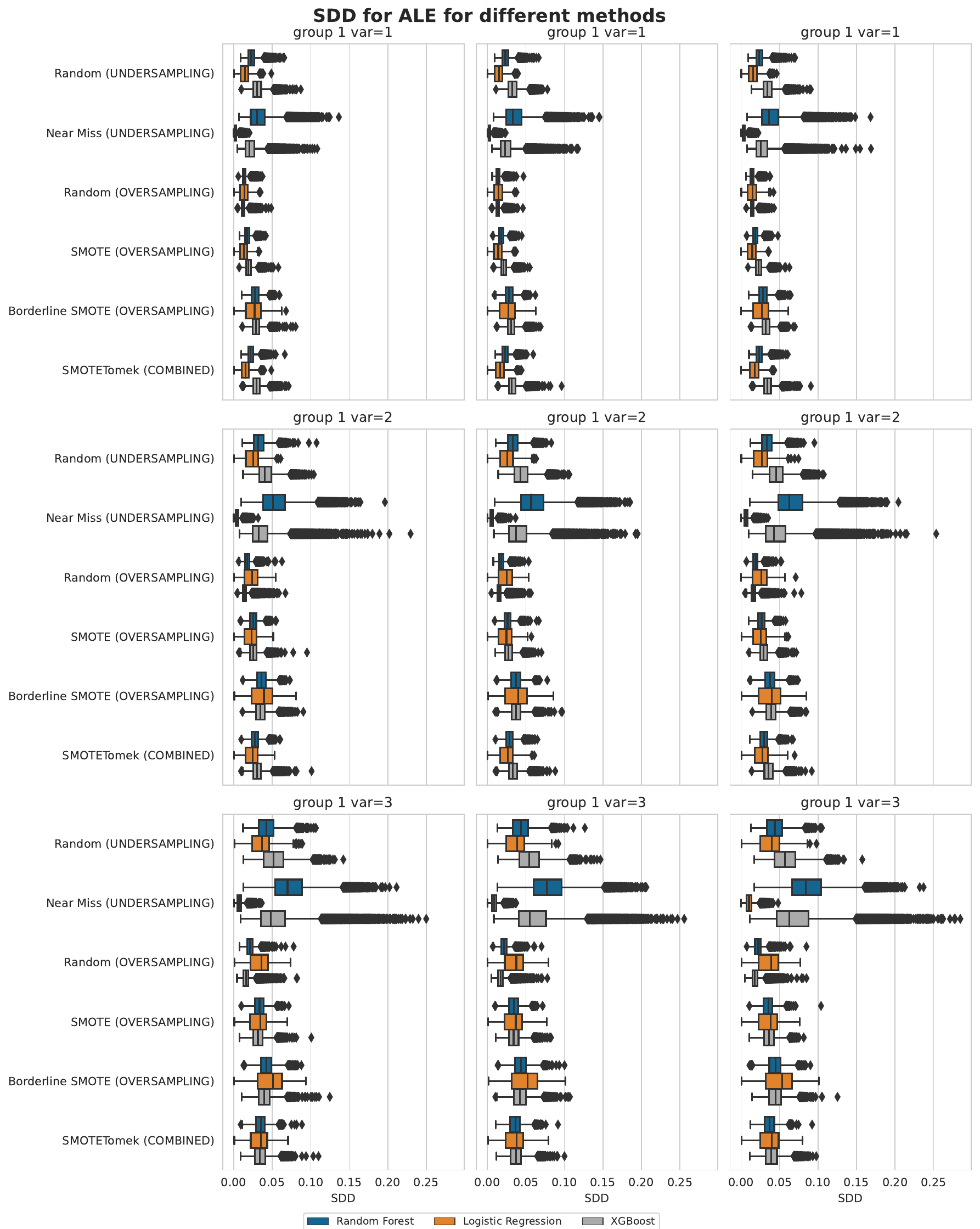}
    \caption{SDD results based on accumulated local effect profiles of the models trained on the simulated dataset. \texttt{group} represents the $\beta_0$ values (group $i$: $\beta_{0i} = \{1.5, 2.5, 3.5, 4.5\}$), and \texttt{var} represents the variance of the error term ($var_j = \{1, 2, 3\}$)}
    \label{fig:truth-ale}
\end{figure}

The SDD values for PDP and ALE profiles between models trained on the simulated datasets are presented in Figure \ref{fig:truth-pdp} and \ref{fig:truth-ale}, respectively. The similarity of SDD values for PDP and ALE indicates that either approach can be used to compare models. The impact of undersampling and other methods on model behavior varies. Undersampling has the smallest effect on the Logistic Regression model but the largest effect on the XGBoost model. Conversely, other methods have a greater impact on the behavior of the logistic regression model and a lesser impact on the random forest model. The effects of these methods become more pronounced as the variance of the model error term and the imbalance ratio of predicted values increase. Thus, the negative impact of balancing methods on model behavior increases with higher variance and imbalance ratios in the model predictions.

\subsection{Benchmark Datasets}
Benchmark datasets are the backbone of large-scale experiments. Their quality of it is very important for generalizing the results obtained in the experiments. \cite{Moniz_and_Cerquiera_2021} and \cite{Singh_and_Vanschoren_2022} proposed benchmark datasets for imbalanced learning. To measure the effect of balancing methods on model behavior in terms of SDD, we propose a new benchmarking set of datasets. For now, SDD works only on continuous variables for now, therefore, we need to create a new imbalanced benchmark dataset.

The proposed benchmarking set of datasets is made up of three main sources: OpenML-100, OpenML-CC18 \cite{OpenML} and the collection of datasets available in \texttt{imblearn} library which was proposed by \cite{dataset2}. The benchmarking set contains only datasets for binary classification tasks that have only continuous columns (categorical and nominal were removed), at least 1000 rows, and an imbalance ratio of at least 1.5. The set is also available via a dedicated class in \texttt{edgaro} package. The list of selected datasets and their details are presented in Table \ref{table:benchmarking-set}. 

\begin{table}
\centering
\caption{Proposed benchmarking set}
\label{table:benchmarking-set}
\begin{tabular}{lrrrr}
\toprule
  \textbf{Dataset name} &  \textbf{IR} &  \textbf{Rows} &  \textbf{Columns} & \textbf{Source} \\
\midrule
          spambase &   1.54 &  4601 &    55 & OpenML-100, OpenML-CC18 \\
    MagicTelescope &   1.84 & 19020 &    10 & OpenML-100\\
steel-plates-fault &   1.88 &  1941 &    13 & OpenML-100, OpenML-CC18\\
       qsar-biodeg &   1.96 &  1055 &    17 & OpenML-100, OpenML-CC18\\
           phoneme &   2.41 &  5404 &     5 & OpenML-100\\
               jm1 &   4.17 & 10880 &    17 & OpenML-100, OpenML-CC18\\
       SpeedDating &   4.63 &  1048 &    18 & OpenML-100\\
               kc1 &   5.47 &  2109 &    17 & OpenML-100, OpenML-CC18\\
             churn &   6.07 &  5000 &     8 & OpenML-CC18\\
               pc4 &   7.19 &  1458 &    12 & OpenML-100, OpenML-CC18\\
               pc3 &   8.77 &  1563 &    14 & OpenML-100, OpenML-CC18\\
           abalone &   9.68 &  4177 &     7 & imblearn\\
          us\_crime &  12.29 &  1994 &   100 & imblearn\\
         yeast\_ml8 &  12.58 &  2417 &   103 & imblearn\\
               pc1 &  13.40 &  1109 &    17 & OpenML-100, OpenML-CC18\\
   ozone-level-8hr &  14.84 &  2534 &    72 & imblearn, OpenML-100, OpenML-CC18 \\
              wilt &  17.54 &  4839 &     5 & OpenML-100, OpenML-CC18\\
      wine\_quality &  25.77 &  4898 &    11 & imblearn\\
         yeast\_me2 &  28.10 &  1484 &     8 & imblearn\\
       mammography &  42.01 & 11183 &     6 & imblearn\\
        abalone\_19 & 129.53 &  4177 &     7 & imblearn\\

\bottomrule
\end{tabular}
\end{table}

\subsection{Real Dataset Experiments}
\begin{figure}
   \begin{subfigure}{0.49\textwidth}
       \centering
       \includegraphics[width=\textwidth,keepaspectratio]{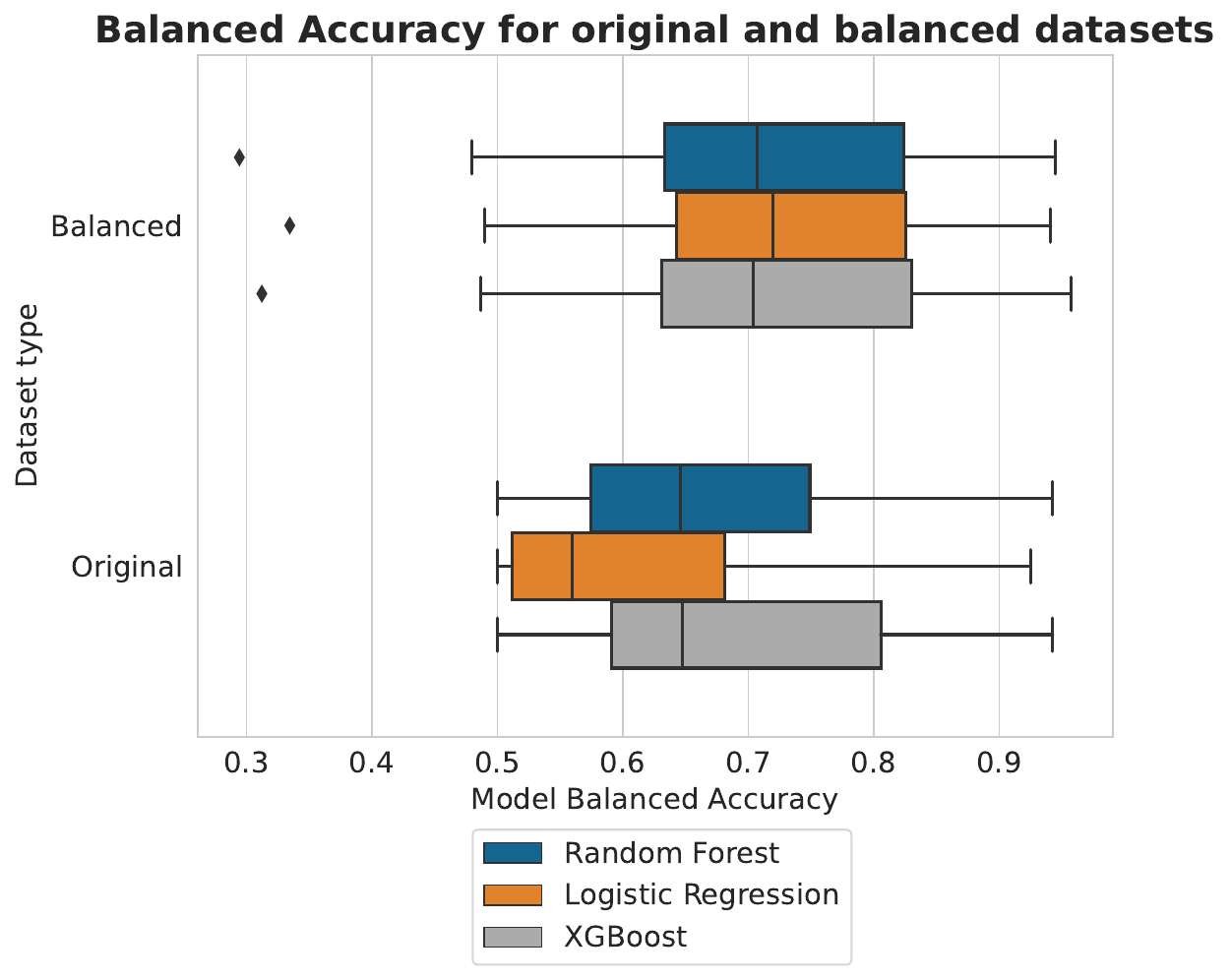}
       \caption{Balanced Accuracy for different models on balanced and original datasets.}
       \label{fig:ba-ds-type}
    \end{subfigure}\hfill
    \begin{subfigure}{0.49\textwidth}
       \centering
       \includegraphics[width=\textwidth,keepaspectratio]{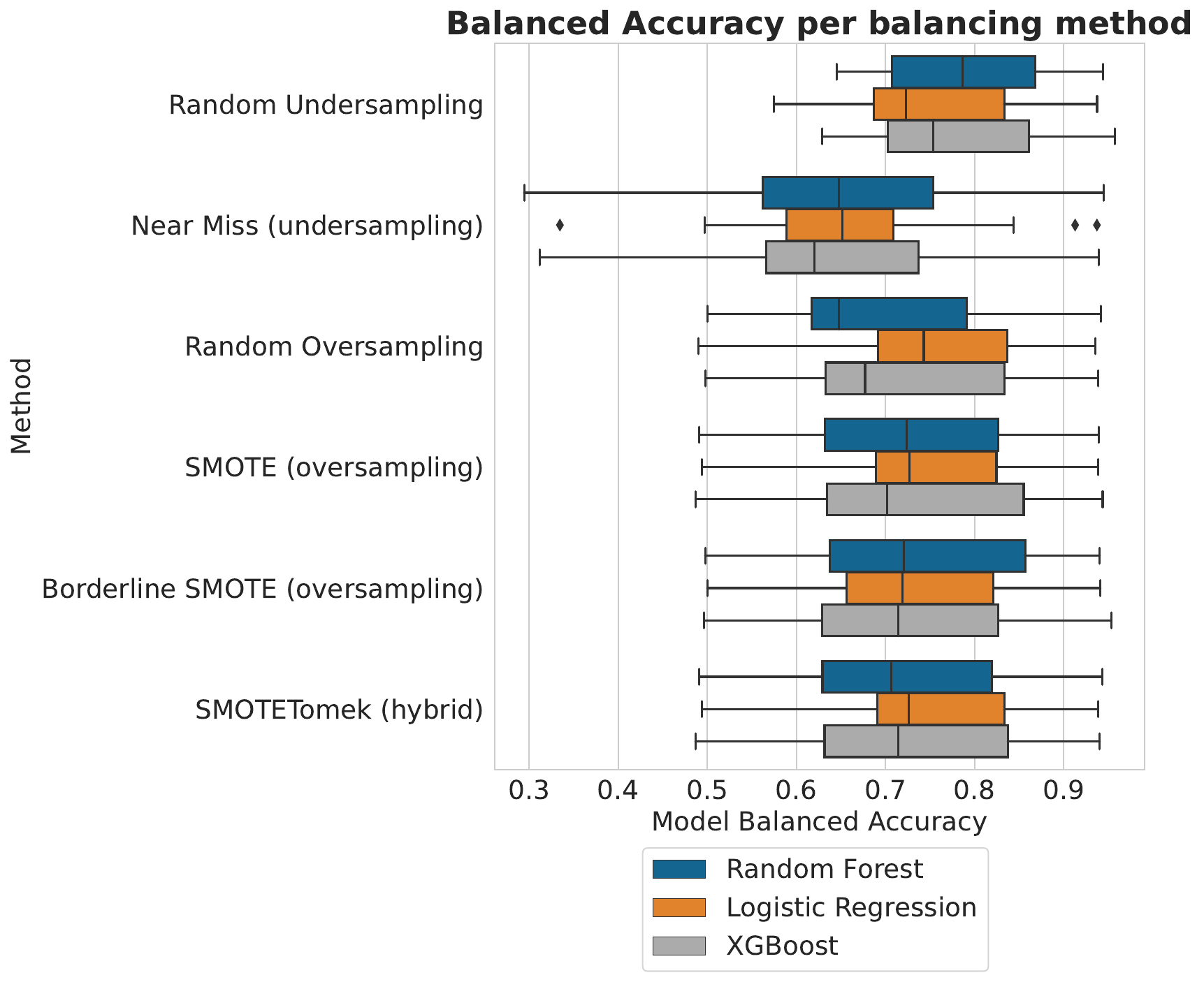}
       \caption{Balanced Accuracy for different methods of balancing datasets.}
       \label{fig:ba-methods}
    \end{subfigure}\hfill
    \caption{Balanced Accuracy results of the real dataset experiments.}
\end{figure}

We ran the experiments on the proposed benchmark dataset. Figure \ref{fig:ba-ds-type} presents the balanced accuracy values for models on both original and balanced datasets. It is clear from the chart that balancing does improve model performance in terms of the evaluation metric. This confirms the results of many studies showing that these methods have a positive influence on the predictive power of the models. The biggest change observed is in the case of XGBoost, which has evolved from the worst to the best model after balancing. Nevertheless, there are some cases where the predictive power of the model decreased significantly. These outliers are the effect of the Near Miss method (Figure \ref{fig:ba-methods}). It is the only one that in some cases prevents the models from learning from the data. Apart from that, Figure \ref{fig:ba-methods} suggests that the best results were obtained after applying the \textbf{Random Undersampling} technique and that the Random Forest model benefited from it the most.

\begin{figure}
    \begin{subfigure}{0.48\textwidth}
       \centering
       \includegraphics[width=\textwidth]{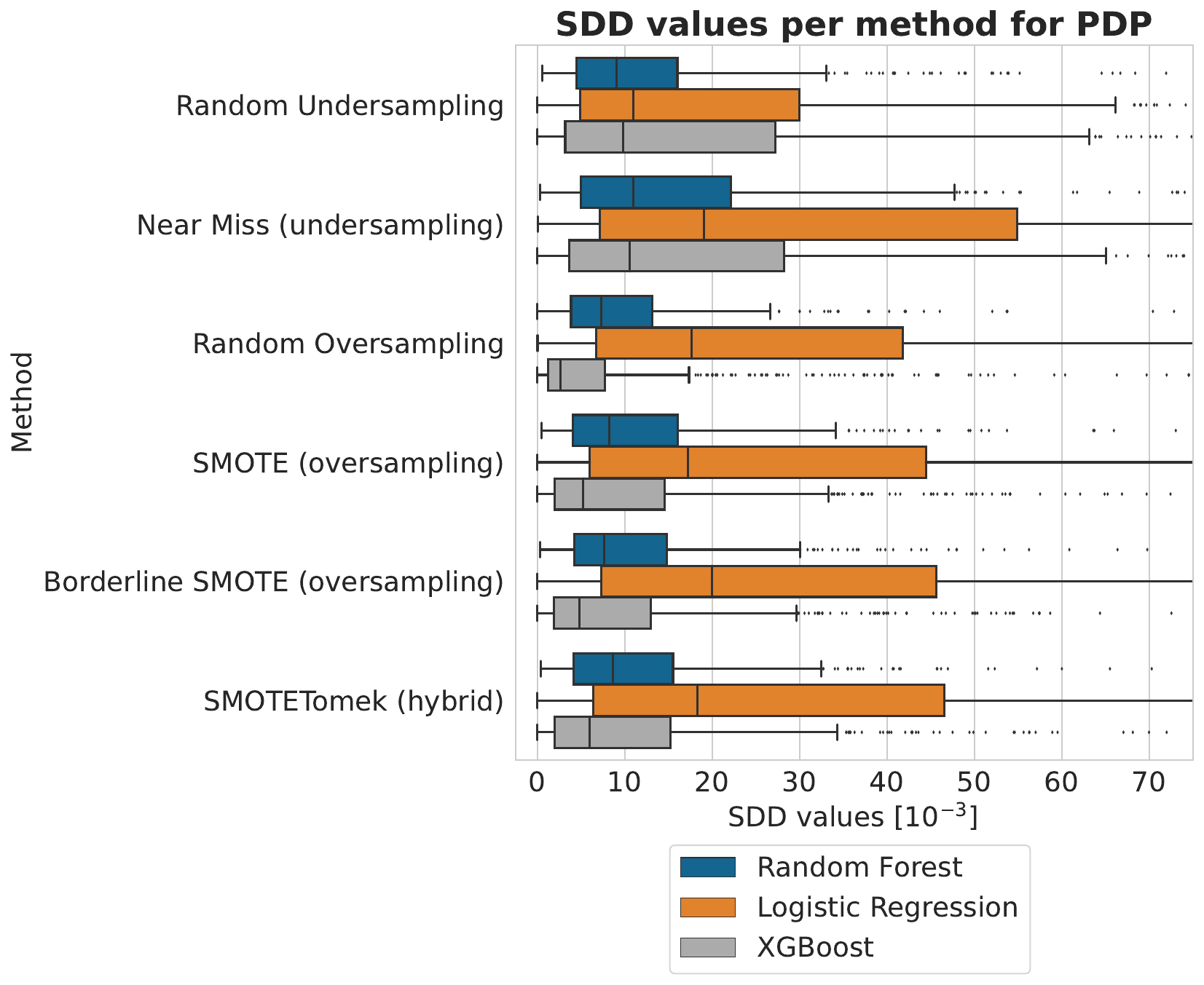}
       \caption{The SDD values per balancing method for Partial Dependence Profiles.}
       \label{fig:sdd-pdp-methods}
    \end{subfigure}\hfill
    \begin{subfigure}{0.48\textwidth}
       \centering
       \includegraphics[width=\textwidth]{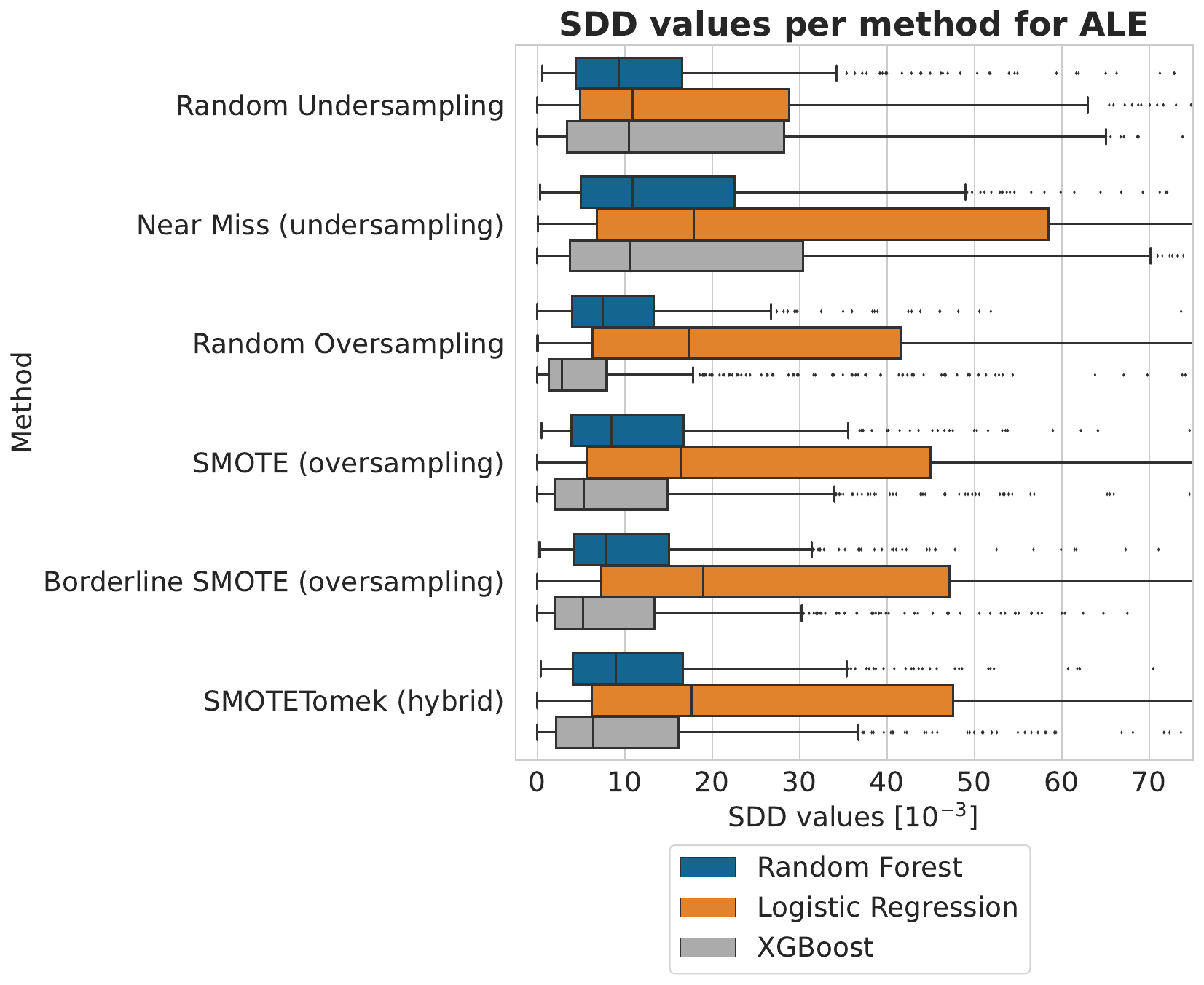}
       \caption{The SDD values per balancing method for Accumulated Local Effects Profiles.}
       \label{fig:sdd-ale-methods}
    \end{subfigure}
    \caption{The SDD results of the real dataset experiments.}
    \label{fig:sdds}
\end{figure}

The SDD values for PDP and ALE profiles between models trained on the original and the balanced datasets are presented in Figure \ref{fig:sdds}. Firstly, the SDD values for PDP and ALE are very alike in these plots. This means that either approach can be used to compare models. Secondly, all balancing methods cause significant changes in the behavior of the Logistic Regression models. Consequently, it can be concluded that this model is biased toward a balanced distribution. Moreover, the Near Miss method, which was the worst in terms of balanced accuracy, also causes the largest changes in SDD values.
On the other hand, the XGBoost models have changed the least after applying Random Oversampling. This means that it is safe to use this technique when training an XGBoost model.

\begin{figure}[htb]
    \centering
    \begin{minipage}{.54\textwidth}
               \includegraphics[width=\linewidth]{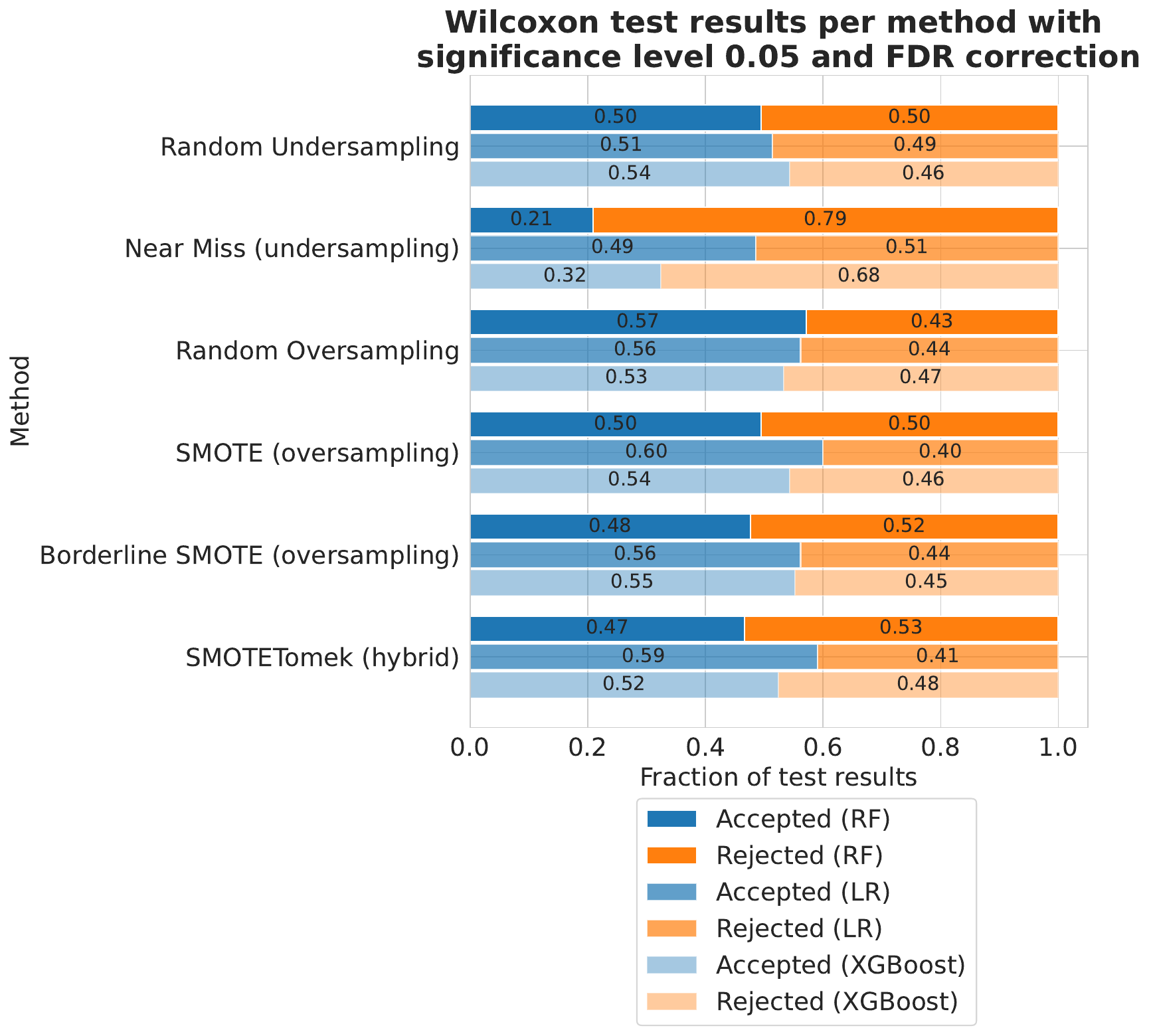}
               \caption{Wilcoxon test results with the significance level of 0.05 and FDR correction.}
               \label{fig:vi-real}
    \end{minipage} \hfill
    \begin{minipage}{0.44\textwidth}
               \includegraphics[width=\linewidth]{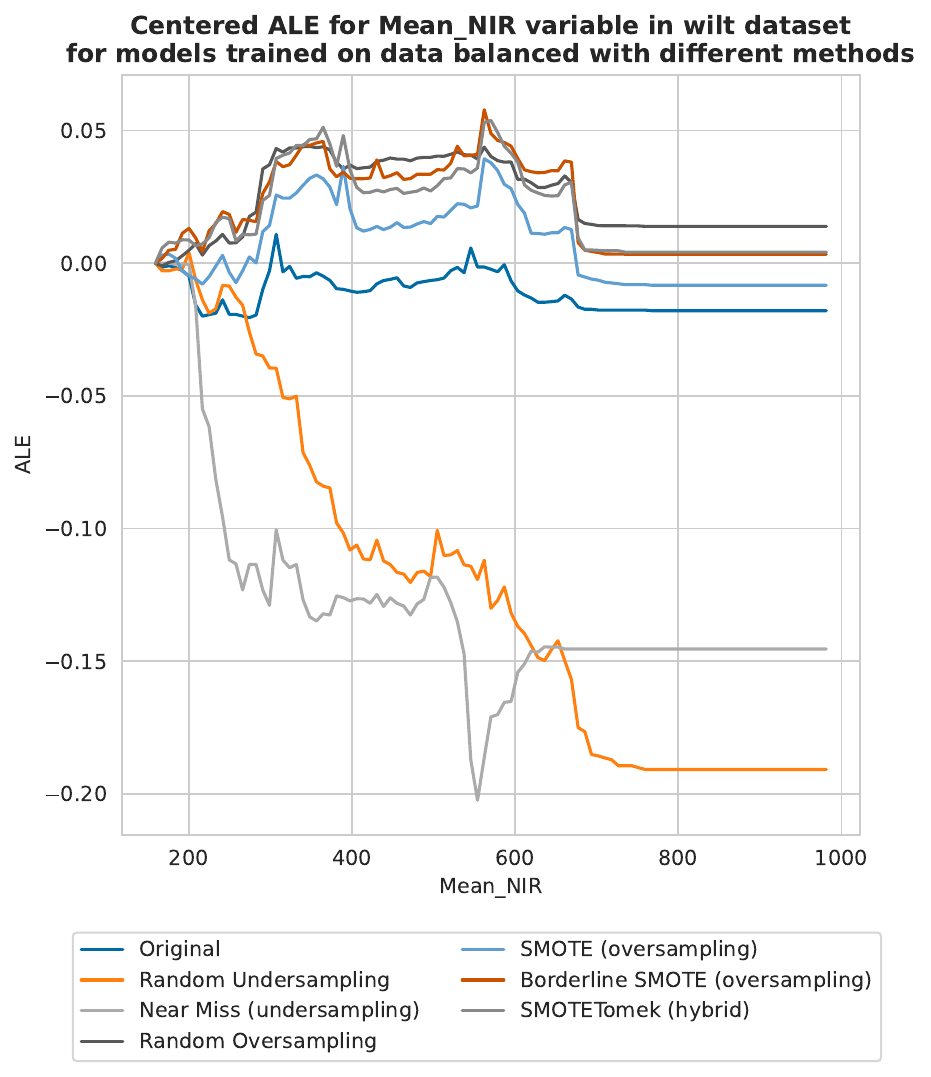}
               \caption{ALE plots for models trained on the wilt dataset balanced with different methods.}
               \label{fig:ale}
    \end{minipage}
\end{figure}

An example of how model behavior can change after balancing is illustrated in Figure \ref{fig:ale}. It presents ALE profiles for Random Forest models trained on the \textit{wilt} dataset. It can be seen that the profiles for Random Undersampling and Near Miss methods have completely different characteristics compared to the original line.

The results of the Wilcoxon test, which compares the Variable Importance profiles of the models trained on the original and balanced datasets, are visualized in Figure \ref{fig:vi-real}. 
This plot confirms earlier observations that the \textbf{Near Miss} method causes the most changes in the model behavior. 
On the other hand, the XGBoost algorithm seems to be the most robust, as it has the largest fraction of accepted tests in all cases.
However, it should be noted that the results obtained by the VI and PDP/ALE methods are not coherent. For example, the smallest SDD values were present for the XGBoost algorithm after applying \textbf{Random Oversampling}. The chart in Figure \ref{fig:vi-real} does not depict that - the larger fraction of the accepted tests has Random Forest.

\subsection{Performance gain plot}

We proposed the performance gain plot that shows the relationship between the models and the balancing methods considered. The performance gain is given in the x-axis in terms of balanced accuracy and the ASDD values, which show the model behavior change based on the PDP or ALE profiles, are given in the y-axis. On such a scatterplot, we can compare two types of changes: in performance and behavior. The higher values on the x-axis and the lower the values on the y-axis is better. Conversely, high model behavior changes and low prediction performance gain. In this direction, we examined the changes in model behavior versus model performance gain by using the performance gain plot on simulated and real datasets.

\begin{figure}
    \centering
    \includegraphics[width=\linewidth]{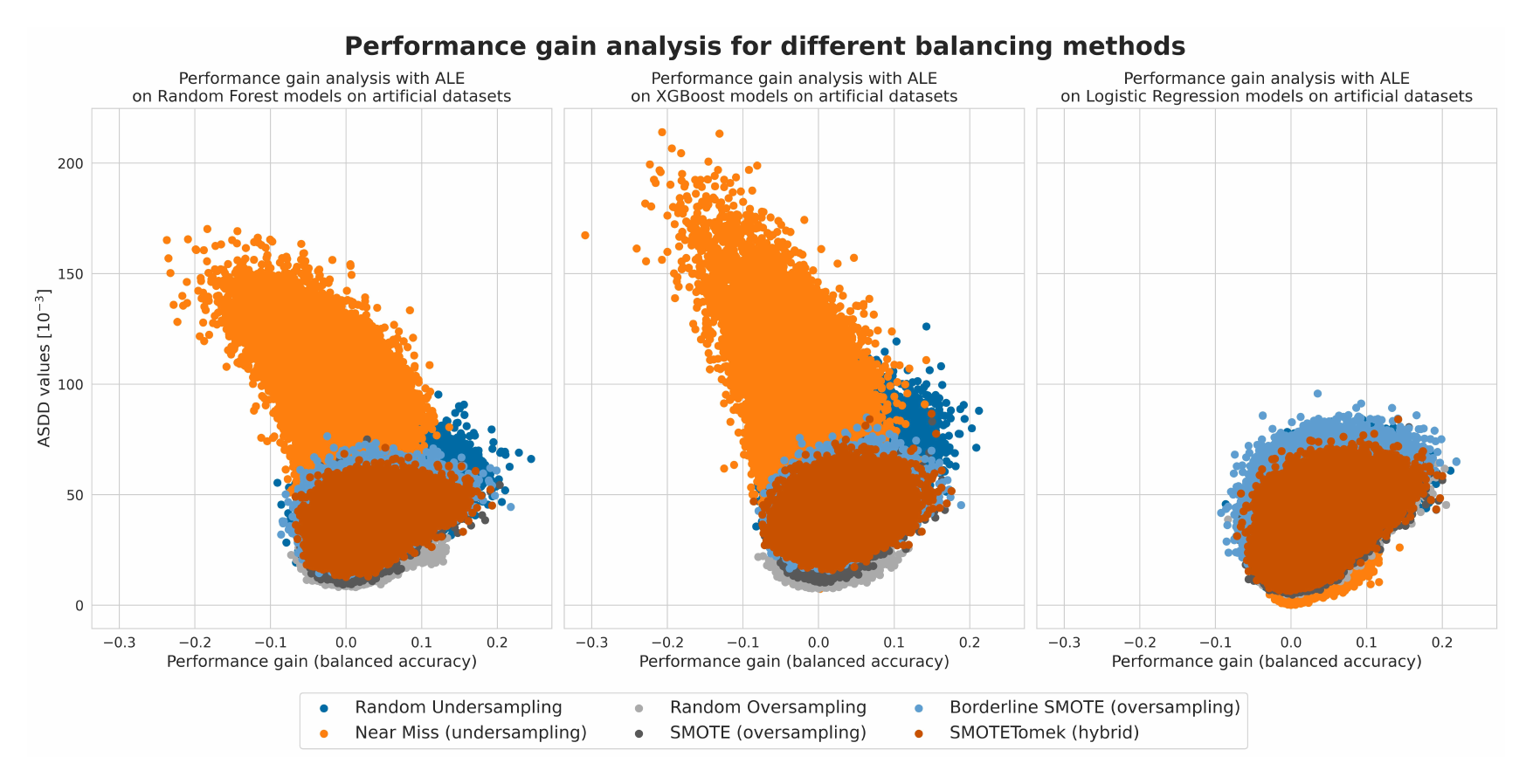}
    \caption{Performance gain plot for simulated datasets.}
    \label{fig:performance_gain_artificial}
\end{figure}

In Figure~\ref{fig:performance_gain_artificial}, the most important observation in the case of Random Forest and XGBoost, the Near Miss method was the one with the highest behavior changes and the largest performance change. On the contrary, the Random Undersampling technique had the highest performance gain, but still at the cost of behavior change. Another conclusion is that the ASDD values tend to be lower for Logistic Regression than for other models. Consequently, it can be concluded that the Near Miss method is the riskier method in terms of model behavior change for Random Forest and XGBoost models.

\begin{figure}
    \centering
    \includegraphics[width=\linewidth]{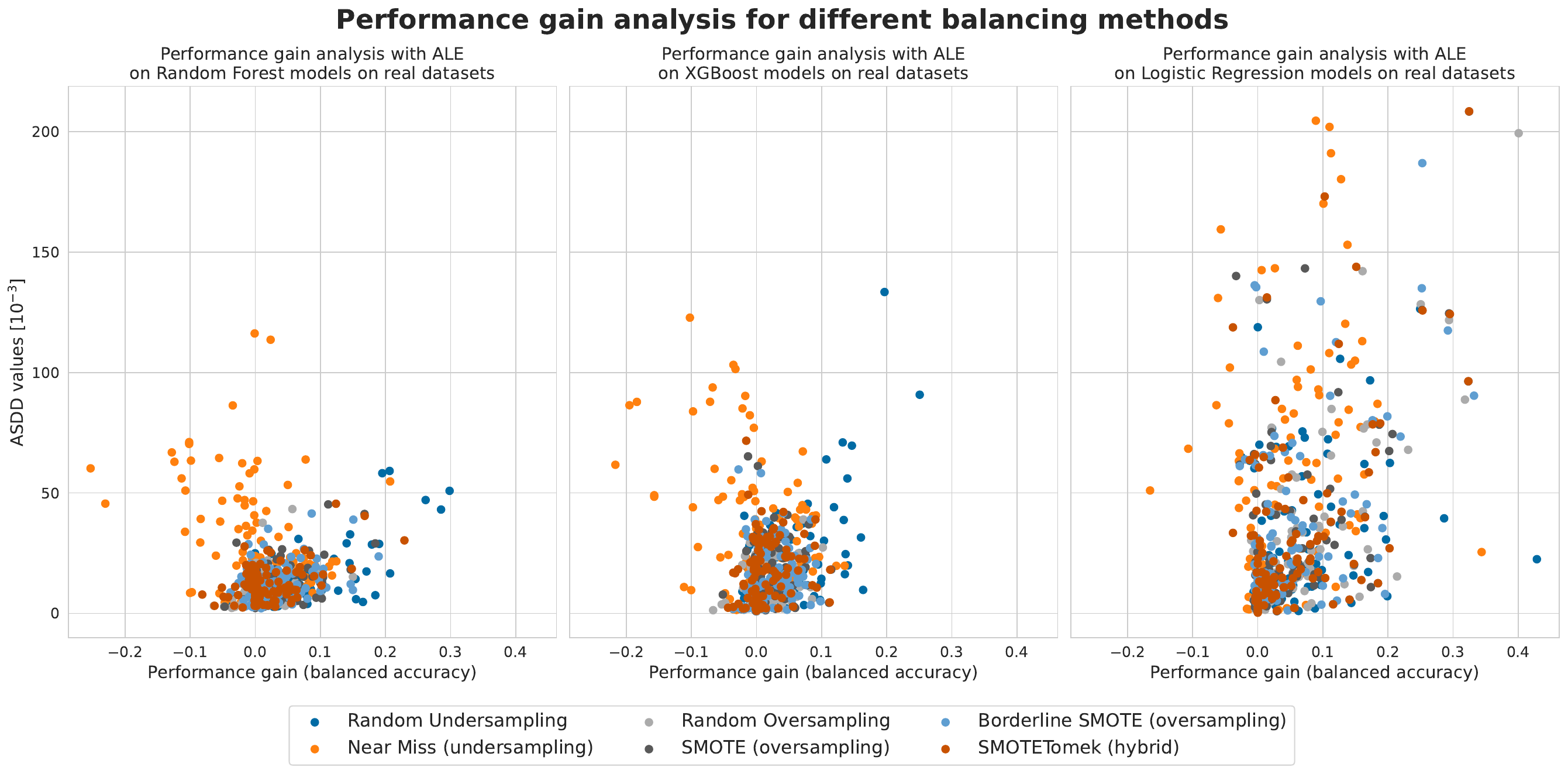}
    \caption{Performance gain plot for real datasets.}
    \label{fig:performance_gain}
\end{figure}

Figure \ref{fig:performance_gain} shows that the most important observation in the case of Random Forest and XGBoost is that for some datasets, the Near Miss method was the one with the highest behavior changes and the largest performance decrease. On the other hand, the Random Undersampling technique had the highest performance gain, but still at the cost of behavior change. Another conclusion is that the ASDD values tend to be higher for Logistic Regression than for other models, while the decrease in the balanced accuracy value is almost not observed (or only to a small extent). Therefore, it can be concluded that data balancing in the case of Logistic Regression is safer in terms of performance, but riskier in terms of behavior change.

\section{Conclusions} \label{sec:conclusions}
In this paper, we investigated the impact of balancing methods on model behavior in imbalanced datasets. We conducted experiments on both simulated and real datasets to measure the impact of different balancing methods on model behavior and performance by using \texttt{edgaro} package. Our results show that \textbf{Random Undersampling} is the most effective method for improving model performance, followed by all \textbf{Oversampling} methods. However, the \textbf{Near Miss} method does not always lead to better performance, especially when the imbalance ratio is high. We also observed that the impact of the balancing methods on model behavior varies depending on the algorithm. These findings are consistent with the results presented in \cite{moniz_and_monteiro_2021} about the \textit{No Free Lunch} concept \cite{schaffer_1994} for imbalanced ML. Thus, we propose to use the performance gain plot to select the optimal balancing method in terms of performance gain and model behavior change. Additionally, we introduced a comprehensive model framework and followed a simulation design similar to previous studies to generate simulated datasets with controlled imbalances. The results of our experiments on these datasets demonstrate that the negative impact of balancing methods on model behavior increases with higher variance and imbalance ratios of model predictions.

In conclusion, our paper provides insights into the trade-offs between model performance and behavior when dealing with imbalanced datasets. Future research can explore alternative balancing methods, such as cost-sensitive learning, or combine multiple methods to further improve model performance and minimize changes in model behavior in imbalanced datasets.

\section*{Supplemental Materials}

The materials for reproducing the experiments performed in Sec~\ref{sec:experiments}, the Python package \href{https://github.com/adrianstando/edgaro}{edgaro}, and the benchmark datasets are accessible at \href{https://github.com/adrianstando/ECML-PKDD-2023-effects-of-data-balancing}{the repository}.

\section*{Acknowledgements} 

The work on this paper was carried out with the support of the Laboratory of Bioinformatics and Computational Genomics and the High-Performance Computing Center of the Faculty of Mathematics and Information Science, Warsaw University of Technology under computational grant number A-22-09. Also, it is financially supported by the NCN Sonata Bis-9 grant 2019/34/E/ST6/00052, and Eskisehir Technical University Scientific Research Projects Commission under grant no. 22ADP367.

\bibliographystyle{unsrt}
\bibliography{references}

\end{document}